\documentclass[runningheads]{llncs}
\usepackage{graphicx}
\usepackage{amsmath}
\usepackage{multirow}
\usepackage{subfigure}
\usepackage{bbding}
\usepackage{booktabs}
\usepackage[normalem]{ulem}
\useunder{ine}{}{}
\usepackage{bm}
\usepackage{wrapfig}
\usepackage{hyperref}
\hypersetup{hidelinks,
	colorlinks=true,
	allcolors=black,
	pdfstartview=Fit,
	breaklinks=true}
\begin{document}
\title{CausE: Towards Causal Knowledge Graph Embedding}

\author{
Yichi Zhang,
Wen Zhang\thanks{Corresponding author.}
}
\institute{Zhejiang University, Hangzhou, China     \\\email{\{zhangyichi2022,zhang.wen\}@zju.edu.cn}}
%
%
%
\maketitle              
\begin{abstract}
Knowledge graph embedding (KGE) focuses on representing the entities and relations of a knowledge graph (KG) into the continuous vector spaces, which can be employed to predict the missing triples to achieve knowledge graph completion (KGC). However, KGE models often only briefly learn structural correlations of triple data and embeddings would be misled by the trivial patterns and noisy links in real-world KGs. To address this issue, we build the new paradigm of KGE in the context of causality and embedding disentanglement. We further propose a \textbf{Caus}ality-enhanced knowledge graph \textbf{E}mbedding (\textbf{CausE}) framework. CausE employs causal intervention to estimate the causal effect of the confounder embeddings and design new training objectives to make stable predictions. Experimental results demonstrate that CausE could outperform the baseline models and achieve state-of-the-art KGC performance. We release our code in \href{https://github.com/zjukg/CausE}{https://github.com/zjukg/CausE}.

\end{abstract}
\section{Introduction}
\label{section::introduction}
Knowledge graphs (KGs) \cite{DBLP:conf/sigmod/BollackerEPST08} modeling the world knowledge with structural triples in the form of \textit{(head entity, relation, tail entity)}, which portrays the relation between the head and tail entity. Expressive KGs have become the new infrastructure of artificial intelligence (AI), which have been widely used in question answering \cite{DBLP:conf/naacl/YasunagaRBLL21}, recommender systems \cite{DBLP:conf/mm/ZhuZZYCZC21}, and fault analysis \cite{chen2022tele}.
\begin{figure}[h]
    \centering
    \includegraphics[width=0.65\linewidth]{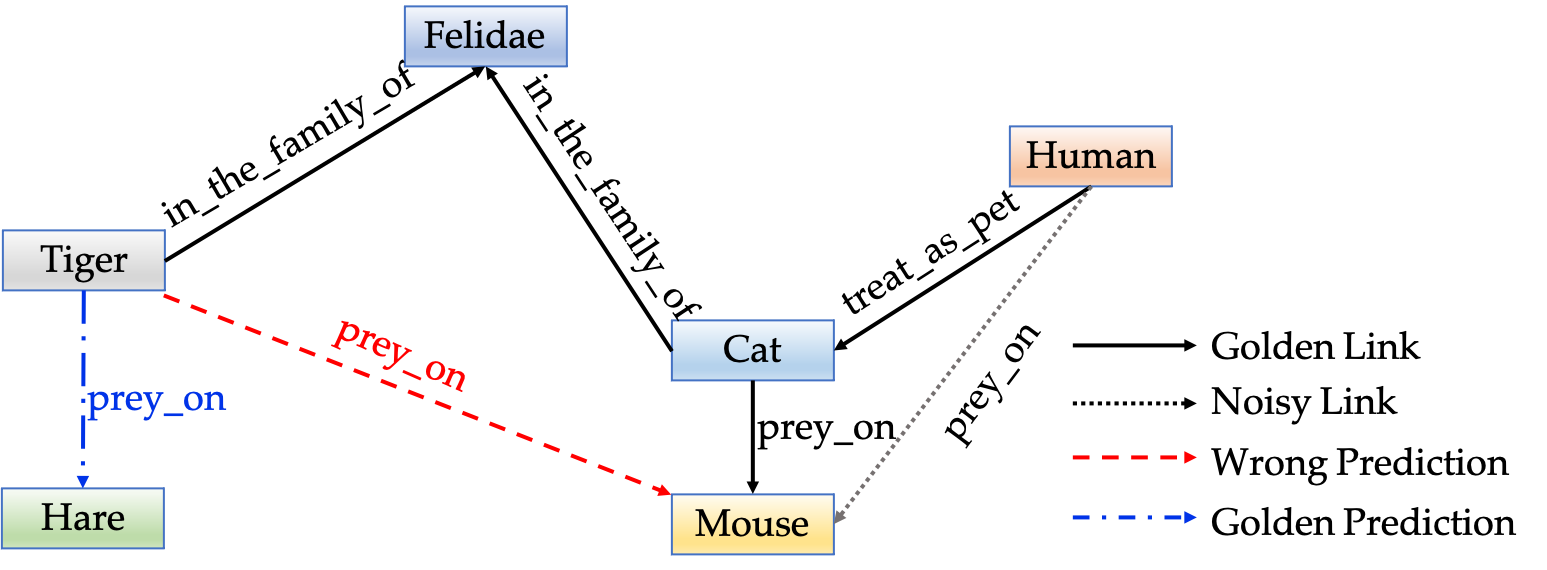}
    \caption{A simple example to explain that the confounding factors like noisy links e.g. (Human, prey\_on, Mouse) and trivial patterns (Both Tiger and Cat are in the family of Felidae) might mislead the link prediction. In this case, the prediction result of (Tiger, prey\_on, ?) would be misled to Mouse.}
    \label{image::case}
\end{figure}
\par KGs are usually inherently incomplete due to their vast diversity and complexity. To address this issue, knowledge graph completion (KGC) has become a popular research topic, aimed at identifying undiscovered triples in KGs. A mainstream solution to KGC is knowledge graph embedding (KGE), which utilizes low-dimensional continuous space to embed entities and relations from the KG. The triple structure is modeled through a score function \cite{DBLP:conf/nips/BordesUGWY13,DBLP:journals/corr/YangYHGD14a,DBLP:conf/iclr/SunDNT19} that measures the plausibility of each triple, forming the basis for predictions in KGC tasks.
\par However, in KGs, various confounding factors (such as trivial structural patterns, noisy links, etc.) may mislead KGE models, resulting in spurious correlations \cite{DBLP:conf/kdd/SuiWWL0C22} being learned and non-causal predictions being made. Figure \ref{image::case} provides an intuitive view of such a situation. While many existing methods propose scoring functions to model different relationship patterns, they overlook the possibility that the knowledge graph data itself may contain information that could mislead the model.
\par To address the mentioned problem, We decouple the embeddings of entities and relations into causal and confounder embeddings. Then we introduce the theory of causal inference \cite{pearl2014interpretation} to model and analyze this problem. We construct the structural causal model (SCM) \cite{pearl2000models} to analyze the KGE task in the context of causality.
Meanwhile, we propose a \textbf{Caus}ality-enhanced knowledge graph \textbf{E}mbedding (CausE) framework to guide the KGE models to learn causal features in the KG. In CausE, we design the intervention operator to implement the backdoor adjustment \cite{pearl2000models}, which would combine the two kinds of embeddings to estimate the effect of the causal and confounder embeddings.
Besides, we design two auxiliary training objectives to enhance the model. We conduct comprehensive experiments on two public benchmarks with the link prediction task to demonstrate the effectiveness of CausE on KGC and make further explorations. The main contribution of this paper can be summarized as follows:
\begin{itemize}
    \item We are the first work to introduce causality theory into the field of KGE.
    \item We propose a new learning paradigm for KGE in the context of causality and design a \textbf{Caus}ality-enhanced knowledge graph \textbf{E}mbedding (\textbf{CausE} for short) framework to learn causal embeddings for KGE models.
    \item We conduct comprehensive experiments on public benchmarks to demonstrate the effectiveness of CausE. We also make
    further exploration to understand it deeply.
\end{itemize}

\section{Related Works}
\subsection{Knowledge Graph Embedding}
\par Knowledge graph embedding \cite{DBLP:journals/tkde/WangMWG17} usually represent the entities and relations of a KG into the low dimensional continuous space to learn the structural features in the KG. A score function is defined in the KGE model to model the triple structure and discriminate the plausibility of triples. 
\par Existing KGE methods \cite{DBLP:conf/nips/BordesUGWY13,DBLP:journals/corr/YangYHGD14a,DBLP:journals/jmlr/TrouillonDGWRB17,DBLP:conf/iclr/SunDNT19,DBLP:conf/aaai/CaoX0CH21,DBLP:conf/acl/ChaoHWC20} 
focus on design elegant and expressive score functions to modeling the triples. 
Translation-based methods \cite{DBLP:conf/nips/BordesUGWY13,DBLP:conf/iclr/SunDNT19,DBLP:conf/acl/ChaoHWC20} modeling the relation as a translation from head to tail in the representation space. TransE \cite{DBLP:conf/nips/BordesUGWY13} treats the translation as a vector addition. RotatE \cite{DBLP:conf/iclr/SunDNT19} represents the relation as a rotation in the complex space. PairRE \cite{DBLP:conf/acl/ChaoHWC20} employs two vectors for relation representation and designs a more complicated score function in the Euclidean space. Besides, semantic matching \cite{DBLP:journals/corr/YangYHGD14a,DBLP:journals/jmlr/TrouillonDGWRB17} models employ latent semantic matching to score the triples, which could be regarded as implicit tensor factorization. DistMult \cite{DBLP:journals/corr/YangYHGD14a} treats the process as 3D tensor factorization and ComplEx \cite{DBLP:journals/jmlr/TrouillonDGWRB17} further extends it to the complex space.
Although various KGE methods are proposed and achieve state-of-the-art knowledge graph completion results, 
no existing methods are concerned with learning the causality of triple structure and making knowledge graph completion better.
\subsection{Causal Inference-Enhanced Graph  Learning}
Causal inference \cite{pearl2000models,pearl2014interpretation} is a popular statistical research topic which aims to discovering causality between data. In recent years, it is becoming increasingly visible to combine causal inference and machine learning to learn the causality from data rather then the correlation for stable and robust prediction. As for graph learning (GL), causal inference also brings a different perspective to the learning paradigm of graphs. CGI \cite{DBLP:conf/sigir/Feng00XWC21} employs causal theory to select trustworthy neighbors for graph convolution networks. CAL \cite{DBLP:conf/kdd/SuiWWL0C22} proposes a causal attention learning framework to learn the causal feature of graphs to enhance the graph classification task. However, there is no existing work to introduce causal theory into the knowledge graph community.

\section{Preliminary}
\label{section::pipeline}
A knowledge graph can be denoted as $\mathcal{G}=(\mathcal{E}, \mathcal{R}, \mathcal{T})$, where $\mathcal{E}$ is the entitiy set, $\mathcal{R}$ is the relation set, and $\mathcal{T}=\{(h,r,t)| h, t \in \mathcal{E}, r \in \mathcal{R}\}$ is the triple set.
\par A KGE model would embed each entity  $e\in\mathcal{E}$ and each relation $r\in \mathcal{R}$ into the continuous vector space and represent each of them with an embedding. We denote $\mathbf{E}^{|\mathcal{E}|\times d_e}$ and $\mathbf{R}^{|\mathcal{R}|\times d_r}$ as the embedding matrix of entity and relation respectively, where $d_e, d_r$ are the dimensions of the entity embeddings and the relation embeddings. Besides, a score function $\mathcal{F}(h, r, t)$ is defined to measure the triple plausibility. The overall target of the KGE model is to give positive triples higher scores and give negative triples lower scores. During training, negative triples are generated by randomly replacing the head or tail entity for positive-negative contrast. We denote the negative triple set as $\mathcal{T}'=\{(h', r, t)| (h, r, t) \in\mathcal{T}, h'\in \mathcal{E}, h'\not=h\}\cup \{(h, r, t')| (h, r, t) \in\mathcal{T}, t'\in \mathcal{E}, t'\not=t\}$. Sigmoid loss proposed by \cite{DBLP:conf/iclr/SunDNT19} is widely used by recent state-of-the-art KGE methods, which could be denoted as:
\begin{equation}
\label{equation::loss}
\begin{aligned}
    \mathcal{L}=\frac{1}{|\mathcal{T}|}\sum_{(h, r, t)\in \mathcal{T}}\Big(-\log\sigma(\gamma-\mathcal{F}(h, r, t))-\sum_{i=1}^{K}p_i\log\sigma(\mathcal{F}(h_i',r_i',t_i')-\gamma)\Big)
\end{aligned}
\end{equation}
where $\sigma$ is the sigmoid function, $\gamma$ is the margin, and $K$ is the number of negative triples generated for each positive triple. The negative triples for $(h, r, t)$ is denoted as $(h_i', r_i', t_i'), i=1,2,\dots, K$. Besides, $p_i$ is the self-adversarial weight \cite{DBLP:conf/iclr/SunDNT19} for each negative triple $(h_i',r_i',t_i')$. It could be denoted as $p_i=\frac{\exp(\alpha\mathcal{F}(h_i',r_i',t_i'))}{\sum_{j=1}^K \exp(\alpha\mathcal{F}(h_j',r_j',t_j'))}$,
where $\alpha$ is the temperature of self-adversarial weight.

\section{Methodology}
In this section, we first present the structural causal model (SCM) for the KGE task. Then we further propose our causality-enhanced KGE framework CausE to learn causal and confounder embeddings with carefully designed objectives.

\subsection{SCM for KGE task}
\begin{wrapfigure}{r}{0.5\textwidth}
    \centering
    \includegraphics[width=\linewidth]{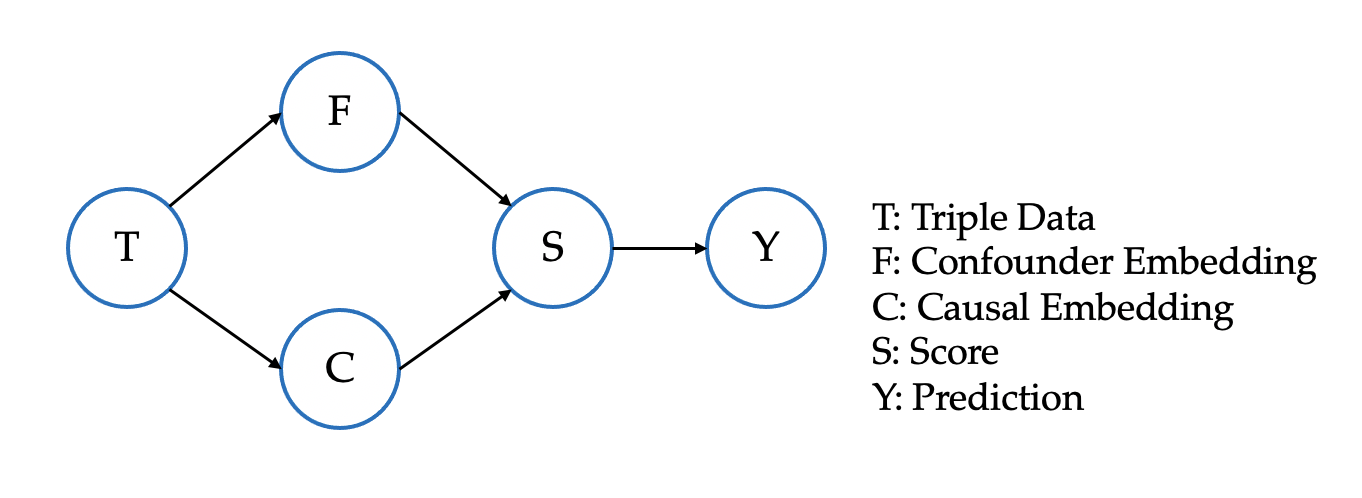}
    \caption{Our SCM for KGE models.}
    \label{image::scm}
\end{wrapfigure}
In KGE models described in Section \ref{section::pipeline}, each entity and relation has a single embedding that encodes both the useful (causal) and harmful (confounder) features.
However, as discussed in Section \ref{section::introduction}, this approach is not robust enough since some local structural information in the KG (e.g. trivial patterns, noisy links) can mislead embedding learning. To develop better embeddings that account for structural causality and make accurate predictions, we introduce the structural causal model (SCM) \cite{pearl2000models} for KGE, as shown in Figure \ref{image::scm}.

\par The SCM defines variables: the triple data $T$, the confounder embeddings $F$, the causal embeddings $C$, the triple score $S$, and the prediction result $Y$. Besides, the SCM demonstrates several causal relations among those variables:
\begin{itemize}
    \item $F\leftarrow T \rightarrow C$. The causal embeddings $C$ encode the implicit knowledge about the triple structure. The confounder embeddings $F$, however, have no contribution to the prediction. As both of them could be learned from the KG data $T$, such causal relations exist in the SCM.
    \item $F\rightarrow S \leftarrow C$. $S$ represents the score of a triple, which is based on both the causal embeddings and confounder embeddings.
    \item $S\rightarrow Y$. We denote $Y$ as the prediction results. The overall target of a KGE model is to predict the proper results $Y$ based on the triple scores $S$ in the inference stage.
\end{itemize}
\par In the original KGE paradigm, the causal and confounder embedding of each entity or relation co-exist in one embedding. With SCM, we explicitly disentangle the structural embeddings from the causal and confounder embeddings and analysis their effects on the prediction results in $Y$. The next question is how to mitigate the impact of $F$ on the final prediction $Y$ to make causal predictions.

\subsection{Causal Intervention}
According to the SCM, both the confounder embeddings $C$ and causal embeddings $F$ could be learned from the triple data, which would be all considered in the triple score $S$. Thus, $F\leftarrow T\rightarrow C\rightarrow S\rightarrow Y$ is a backdoor path \cite{pearl2014interpretation} and $F$ is the confounder between $C$ and $Y$. 
\par To make causal predictions based on causal embeddings $C$, we need to model $P(Y|C)$. However, the backdoor path creates a confounding effect of $F$ on the probability distribution $P(Y|C)$, opening a backdoor from $F$ to $Y$. Therefore, it is crucial to block the backdoor path and reduce the impact of confounder embeddings. This will allow KGE models to make predictions by utilizing the causal embeddings fully. Causality theory \cite{pearl2000models,pearl2014interpretation} provides powerful tools to solve the backdoor path problem. 
\par We employ do-calculus \cite{pearl2000models,pearl2014interpretation} to make the causal intervention on the variable $C$, which could \textbf{cut off the backdoor path} $F\leftarrow T\rightarrow C\rightarrow S\rightarrow Y$. With the help of do-calculus, the influence from the confounder $F$ to $C$ is manually cut off, which means $C, F$ are independent. Our target turns to estimate $P(Y|do(C))$ instead of the confounded $P(Y|C)$. Combined with Bayes Rule and the causal assumptions \cite{pearl2000models,pearl2014interpretation}, we could deduce as follows:
\begin{equation}
    \label{equation::backdoor}
    \begin{aligned}
        P(Y|do(C))=P(Y|S)\sum\limits_{d \in \mathcal{D}}P(S|C, d)P(d)
    \end{aligned}
\end{equation}
\par The above derivation shows that to estimate the causal effect of $C$ on $Y$, it is necessary to consider the scores with both causal and counfounder embeddings. This can be understood as re-coupling the decoupled embeddings and using them to calculate the score of the triple.
In the next section, we would propose our \textbf{Caus}ality-enhanced knowledge graph \textbf{E}mbedding (CausE) framework and implement the backdoor adjustments mentioned above.

\subsection{CausE Framework}

In this section, we would demonstrate our \textbf{Caus}ality-enhanced knowledge graph \textbf{E}mbedding (CausE) framework. We would first describe the basic settings of CausE and emphasize how we implement the backdoor adjustment in the CausE.

\subsubsection{Basic Definition}
\begin{figure}[]
    \centering
    \includegraphics[width=0.8\linewidth]{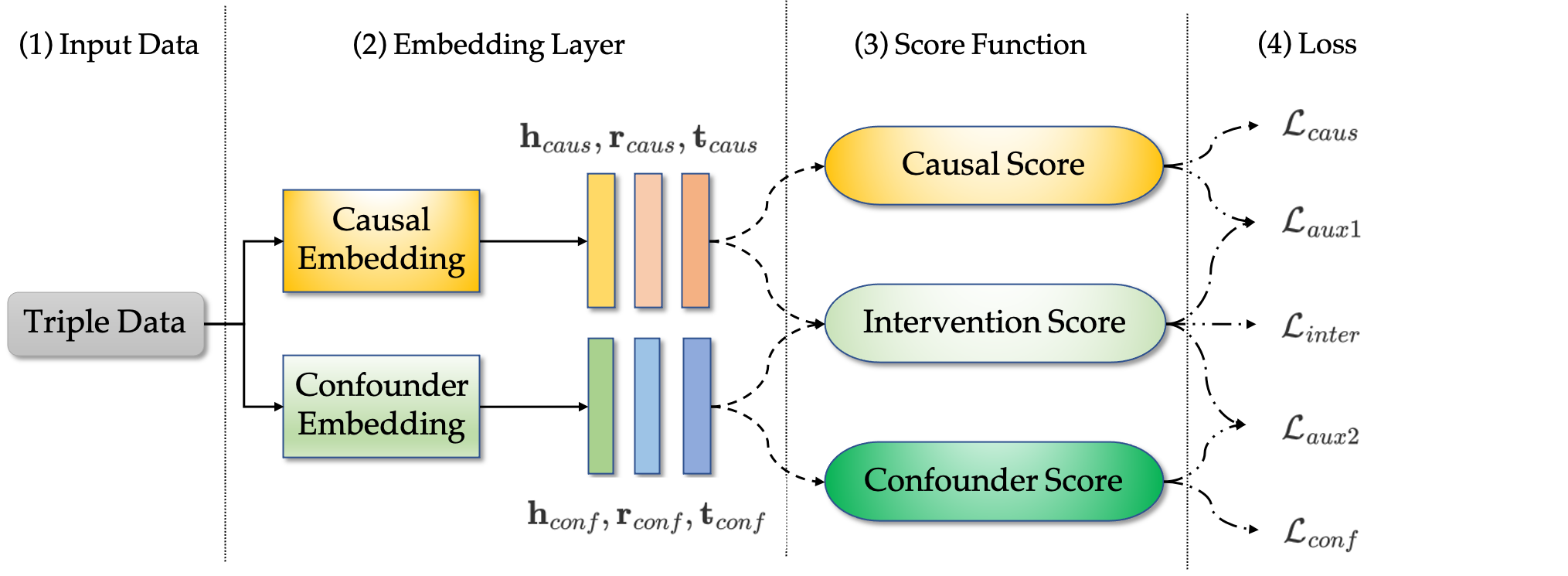}
    \caption{The overall architecture of CausE. We disentangle the embeddings into two parts called causal and confounder embeddings respectively while applying three score functions. We also design five loss functions to train these embeddings while the causal intervention is integrated into them.}
    \label{image::cause}
\end{figure}

The overall framework of CausE is shown in Figure \ref{image::cause}. In the embedding layer, we define two embeddings called causal embedding and confounder embedding for each entity and relation in the KG, aiming to achieve the disentanglement of causal and confounder features.
Specifically, for each entity $e\in \mathcal{E}$, we define a causal embedding $\bm{e}_{caus}$ and a confounder embedding $\bm{e}_{conf}$ for it. Similarly, for each relation $r\in\mathcal{R}$, the two embeddings are $\bm{r}_{caus}$ and $\bm{r}_{conf}$. Such design is consistent with the SCM in Figure \ref{image::scm}.
\par As for the score function, we employ three score functions $\mathcal{F}_{caus}, \mathcal{F}_{conf}, \mathcal{F}_{inter}$, which are called causal score, confounder score, and intervention score respectively. The three score functions are in the same form but can be any general score functions proposed by the existing KGE models. Besides, we design several loss functions to guide the training process of CausE. We would describe the details of the score functions and their corresponding loss functions.

\subsubsection{Causal and Confounder Scores}
\par The causal score function $\mathcal{F}_{caus}(h, r, t)$ takes the causal embeddings $\bm{h}_{caus}, \bm{r}_{caus}, \bm{t}_{caus}$ of $h, r, t$ as input and calculate the causal score the triple. According to our assumption, the causal embeddings are expected to make reasonable and causal predictions. Thus, the causal score $\mathcal{F}_{caus}(h, r, t)$ should still follow the general rule of KGE models: positive triple should have higher scores. We apply sigmoid loss function with self-adversarial negative sampling as the loss function to train the causal embeddings. The causal loss $\mathcal{L}_{caus}$ has the same form as Equation \ref{equation::loss}, which is based on $\mathcal{F}_{caus}$.

\par Meanwhile, the confounder score function $\mathcal{F}_{conf}(h, r, t)$ would calculate the confounder score of the confounder embeddings $\bm{h}_{conf}$, $\bm{r}_{conf}$, $\bm{t}_{conf}$. Different from the causal embeddings, we assume that confounder embeddings learn the harmful features from the KGs and they make no positive contribution to the reasonable prediction. Hence, the confounder score $\mathcal{F}_{caus}(h, r, t)$ should be close to the confounder score of negative triples, which means the KGE model is misled by the harmful features and could not distinguish the positive triple from high plausibility from the negative triples. Therefore, we apply the mean squared error (MSE) loss to train the confounder embeddings. The training objective can be denoted as:
\begin{equation}
    \begin{aligned}
        \mathcal{L}_{conf}=\frac{1}{|\mathcal{T}|}\sum_{(h, r, t)\in \mathcal{T}} \Big(\mathcal{F}_{caus}(h, r, t)-\sum_{i=1}^{K}p_i\mathcal{F}_{caus}(h_i',r_i',t_i')\Big) ^2
    \end{aligned}
\end{equation}
\par 
By the two loss functions proposed above, we could achieve the disentanglement of the causal and confounder embeddings. 
\subsubsection{Intervention Scores}
\label{section::intervention}
As shown in Equation \ref{equation::backdoor}, we need to implement the backdoor adjustment. As we mentioned above, the formula for backdoor adjustment can be understood as jointly measuring the triple plausibility with both causal and confounder embeddings, while considering all possible confounder embedding. This is equivalent to recombining the two decoupled embeddings into the original embeddings and computing the score.
\par We call this score an intervention score $\mathcal{F}_{inter}(h, r, t)$. Besides, we propose a \textbf{intervetion operator} $\Phi$ to recombine the two embeddings and get the intervention embeddings as the output. This process can be denoted as:

\begin{equation}
\bm{e}_{inter}=\Phi(\bm{e}_{caus},\bm{e}_{conf}),\bm{e}\in\{\bm{h}, \bm{t}\}\quad \bm{r}_{inter}=\Phi(\bm{r}_{caus},\bm{r}_{conf})
\end{equation}
We employ the addition operation as the intervention operation. Hence, we could calculate the intervetion score $\mathcal{F}_{inter}(h, r, t)$ with the intervention embedding $\bm{h}_{inter}, \bm{r}_{inter}, \bm{t}_{inter}$. From another perspective, causal intervention is such a process that employs the confounder embeddings to disrupt the prediction of the causal embeddings to estimate the causal effect of the confounder embeddings. We expect the intervention scores could still lead to reasonable predictions. Thus, the training objective $\mathcal{L}_{inter}$ is also a sigmoid loss like \ref{equation::loss} based on $\mathcal{F}_{inter}$.

\subsubsection{Auxiliary Objectives}
To further improve the performance of CausE, we utilize the intervention score and propose two auxiliary training objectives.
\par As we mentioned above, the intervention embeddings can be regarded as the causal embeddings perturbed by the confounder embeddings. Therefore, the effectiveness of the causal scores should be worse than the causal scores but better than the confounder scores. Based on such an assumption, we design two auxiliary training objectives. The first auxiliary objective is between the causal and intervention scores. We apply the sigmoid loss function to make the contrast between them and push the causal scores higher than the intervention scores:
\begin{equation}
\begin{aligned}
    \mathcal{L}_{aux1}=\frac{1}{|\mathcal{T}|}\sum_{(h, r, t)\in \mathcal{T}}\Big(-\log\sigma(\gamma-\mathcal{F}_{caus}(h, r, t))-\log\sigma(\mathcal{F}_{inter}(h, r, t)-\gamma)\Big)
\end{aligned}
\end{equation}
\par The second auxiliary objective $\mathcal{L}_{aux2}$ is similarly designed as $\mathcal{L}_{aux1}$ to push the intervention scores higher than the confounder scores. In summary, the overall training objective of CausE is:
\begin{equation}
\mathcal{L}=\mathcal{L}_{caus}+\mathcal{L}_{conf}+\mathcal{L}_{inter}+\mathcal{L}_{aux1}+\mathcal{L}_{aux2}
\end{equation}
\section{Experiments}
In this section, we will demonstrate the effectiveness of our methods with comprehensive experiments. We first detailedly introduce our experimental settings in Section \ref{content::settings}. Then we would demonstrate our results to answer the following questions:
\begin{itemize}
    \item \textbf{RQ1}: Could CausE outperform the existing baseline methods in the knowledge graph completion task?
    \item \textbf{RQ2}: How does CausE perform in the noisy KGs?
    \item \textbf{RQ3}: How much does each module of CausE contribute to the performance?
    \item \textbf{RQ4}: Do the learned embeddings achieve our intended goal?
\end{itemize}
\subsection{Experiment Settings}
\label{content::settings}
\subsubsection{Datasets / Tasks / Evaluation Protocols.}

In the experiments, we use two benchmark datasets FB15K-237 \cite{DBLP:conf/emnlp/ToutanovaCPPCG15} and WN18RR \cite{DBLP:conf/aaai/DettmersMS018}. 

\label{section::evalutation}
We evaluate our method with link prediction task, which is the main task of KGC. Link prediction task aims to predict the missing entities for the given query $(h, r, ?)$ or $(?, r, t)$. We evaluate our method with mean reciprocal rank (MRR), and Hit@K (K=1,3,10) following \cite{DBLP:conf/iclr/SunDNT19}. Besides, we follow the filter setting \cite{DBLP:conf/nips/BordesUGWY13} which would remove the candidate triples that have already appeared in the training data to avoid their interference.
\subsubsection{Baselines.}
As for the link prediction task, we select several state-of-the-art KGE methods, including translation-based methods (TransE \cite{DBLP:conf/nips/BordesUGWY13}, RotatE \cite{DBLP:conf/iclr/SunDNT19}, PairRE \cite{DBLP:conf/acl/ChaoHWC20}), semantic matching methods (DistMult \cite{DBLP:journals/corr/YangYHGD14a}, ComplEx \cite{DBLP:journals/jmlr/TrouillonDGWRB17}), quaternion-based methods (QuatE \cite{DBLP:conf/nips/0007TYL19}, DualE \cite{DBLP:conf/aaai/CaoX0CH21}), and neural network based methods (ConvE \cite{DBLP:conf/aaai/DettmersMS018}, MurP \cite{DBLP:conf/nips/BalazevicAH19}). We report the baseline results from the original paper.

\subsubsection{Parameter Settings.}
We implement CausE framework to five representative score functions: TransE \cite{DBLP:conf/nips/BordesUGWY13}, DistMult \cite{DBLP:conf/aaai/WangZFC14}, ComplEx \cite{DBLP:journals/jmlr/TrouillonDGWRB17}, PairRE \cite{DBLP:conf/acl/ChaoHWC20}, and DualE \cite{DBLP:conf/aaai/CaoX0CH21}. We apply grid search to tune the hyper-parameters to find the best results of CausE. We search the embedding dimension of the KGE model $d_e, d_r \in \{256, 512, 1024\}$, the margin $\gamma\in\{0, 4, 6, 8\}$, the training batch size $\in \{512, 1024\}$, the temperature $\alpha \in \{1.0, 2.0\}$, the negative sample number $N_k\in\{64, 128, 256\}$, and the learning rate $\eta \in$$ \{1e^{-3}, 1e^{-4}, 2e^{-5}\}$. We conduct all the experiments on Nvidia GeForce 3090 GPUs with 24 GB RAM.

\subsection{Main Results (RQ1)}

\begin{table*}[]
\caption{Link prediction results on FB15K-237 and WN18RR. The best results are \textbf{bold} and the second best results are \underline{underlined} for each metrics.}
\label{table::linkprediction}
\centering
\resizebox{0.75\columnwidth}{!}{
\centering
\begin{tabular}{c|cccc|cccc}
\toprule
\multirow{2}{*}{Model} & \multicolumn{4}{c|}{FB15K-237}                                    & \multicolumn{4}{c}{WN18RR}                                        \\
                       & MRR            & Hit@10         & Hit@3          & Hit@1          & MRR            & Hit@10         & Hit@3          & Hit@1          \\ \midrule
TransE \cite{DBLP:conf/nips/BordesUGWY13}                 & 0.279          & 0.441          & 0.376          & 0.198          & 0.224          & 0.520          & 0.390          & 0.022          \\
DistMult \cite{DBLP:journals/corr/YangYHGD14a}               & 0.281          & 0.446          & 0.301          & 0.199          & 0.444          & 0.504          & 0.470          & 0.412          \\
ComplEx \cite{DBLP:journals/jmlr/TrouillonDGWRB17}               & 0.278          & 0.450          & 0.297          & 0.194          & 0.449          & 0.530          & 0.469          & 0.409          \\
ConvE \cite{DBLP:conf/aaai/DettmersMS018}                 & 0.312          & 0.497          & 0.341          & 0.225          & 0.456          & 0.531          & 0.470          & 0.419          \\
RotatE \cite{DBLP:conf/iclr/SunDNT19}                 & 0.338          & 0.533          & 0.375          & 0.241          & 0.476          & 0.571          & 0.492          & 0.428          \\
MurP \cite{DBLP:conf/nips/BalazevicAH19}                  & 0.336          & 0.521          & 0.370          & 0.245          & 0.475          & 0.554          & 0.487          & 0.436          \\
QuatE \cite{DBLP:conf/nips/0007TYL19}                  & 0.311          & 0.495          & 0.342          & 0.221          & 0.481          & \textbf{0.564}          & 0.500          & 0.436          \\
DualE \cite{DBLP:conf/aaai/CaoX0CH21}                  & 0.330          & 0.518          & 0.363          & 0.237          & {\underline{0.482} }    & { 0.561} & \underline{ 0.500}    & \underline{ 0.440}    \\
PairRE \cite{DBLP:conf/acl/ChaoHWC20}                 & \underline{ 0.351}    & \underline{ 0.544}    & \underline{ 0.387}    & \underline{ 0.256}    & -              & -              & -              & -              \\ \midrule
CausE (TransE)                  & 0.332 & 0.517 & 0.368 & 0.234 & 0.227 & 0.536 & 0.391 & 0.023
\\
CausE (DistMult)                  & 0.298 & 0.473 & 0.327 & 0.212 & 0.447	&	0.517 &	0.452 &	0.415
\\
CausE (ComplEx)               & 0.324	&	0.504	& 0.357	& 0.234 & 0.467	&	0.527	& 0.482 &	0.436
\\
CausE (SOTA)                  & \textbf{0.355} & \textbf{0.547} & \textbf{0.392} & \textbf{0.259} & \textbf{0.486} & \underline{0.562}    & \textbf{0.502} & \textbf{0.446}
\\
\bottomrule
\end{tabular}}
\end{table*}

Our main experiment results are in Table \ref{table::linkprediction}. From the results, we could find that 
The CausE could outperform the baseline methods on the two benchmarks. For example, CausE can achieve a relatively 1.4\% Hit@1 improvement on the WN18RR dataset. Such results demonstrate that CausE becomes a new state-of-the-art KGE method.
\par Meanwhile, CausE is a universal framework and can be applied in various KGE models. The results in Table \ref{table::linkprediction} also demonstrate that CausE could enhance the performance of various KGE models, compared with the corresponding baselines trained w/o CausE. For example, the MRR results on the FB15K-237 dataset of the TransE/DistMult/ComplEx models get relative improvement by 18.9\%, 13.5\%, and 16.5\% respectively. We speculate that this is due to the design defects in the early KGE models, which would mislead the model to learn the confounder features in the KG and make non-causal predictions in the inference stage. Overall, we show that CausE can outperform the baseline methods in various score functions. Thus, the \textbf{RQ1} is solved.

\subsection{Link Prediction on Noisy KG (RQ2)}
\par
\begin{figure}[h]
  \centering
  \subfigure[ComplEx + FB15K-237]{\includegraphics[width=0.48\linewidth]{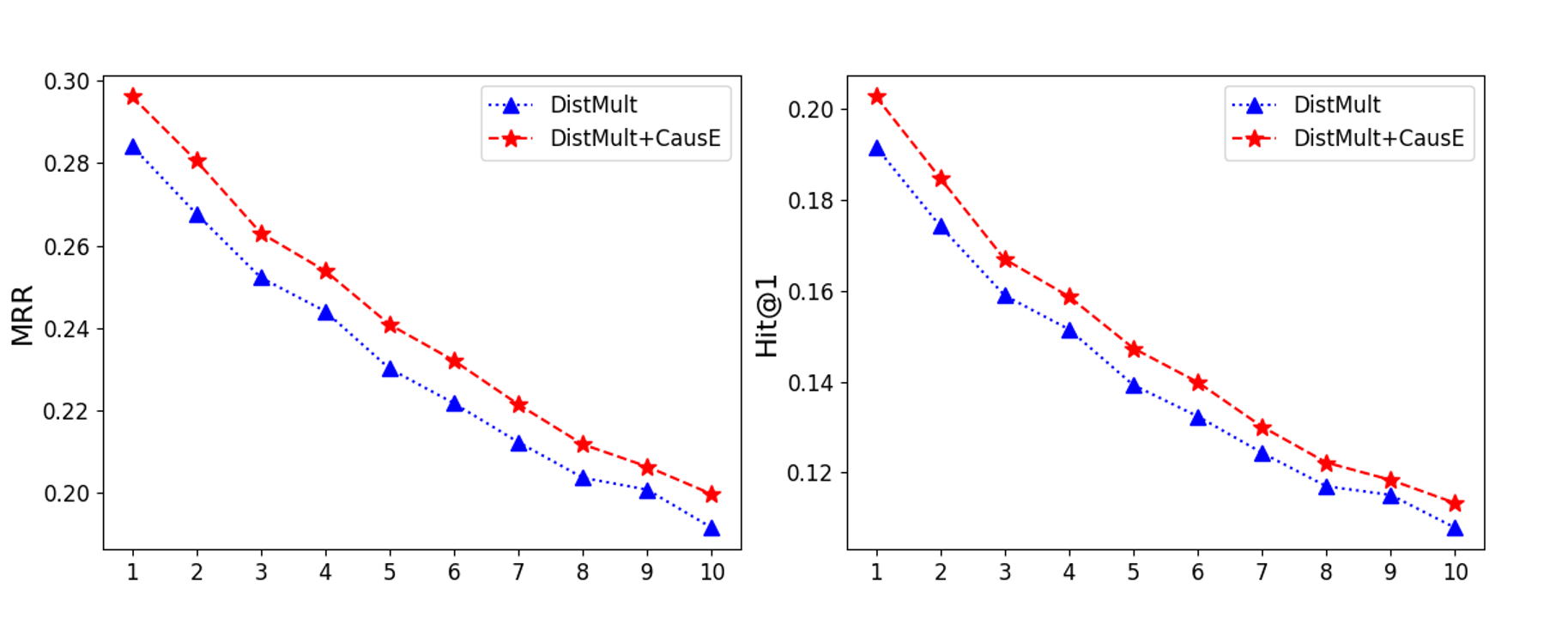}}
  \subfigure[DistMult + WN18RR]{\includegraphics[width=0.48\linewidth]{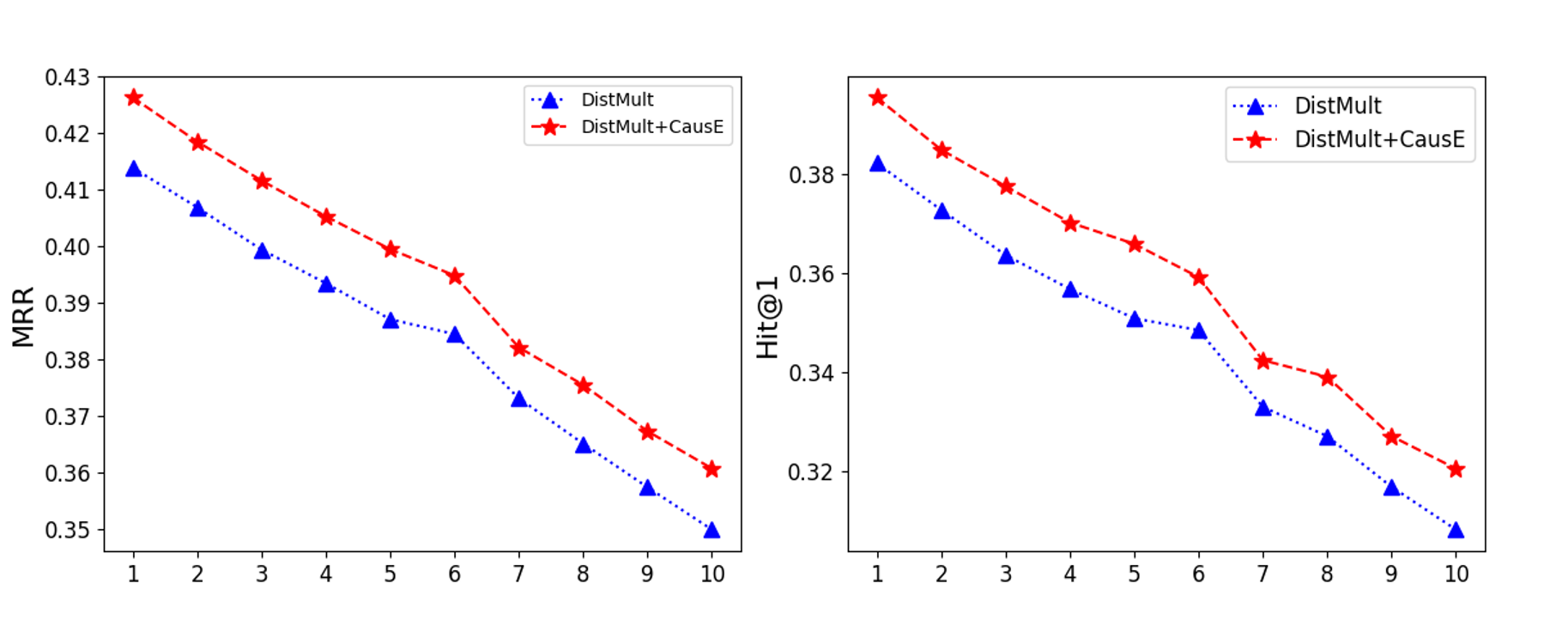}}
  
  \caption{The noisy link prediction results. We report the Hit@1 and MRR results for different experiment settings. The x-axis represents the noisy rate (\%) of the training dataset.}
  \label{img::noisy}
\end{figure}

 To answer the \textbf{RQ2}, we make further exploration on the noisy link prediction task, aiming to validate the robustness of CausE on noisy KGs. We set a parameter called noisy rate $\lambda$, it is defined as $\lambda=\frac{|\mathcal{T}_{noisy}|}{|\mathcal{T}_{train}|}$,
where  $\mathcal{T}_{noisy}\subset \mathcal{T}_{train}$
is the noisy link set of the training set. We
generate noisy KGs by randomly replacing the positive triples and setting the noisy rate $\lambda$ from 1\% to 10\%. We conduct experiments on these noisy datasets with DistMult \cite{DBLP:journals/corr/YangYHGD14a} and ComplEx \cite{DBLP:journals/jmlr/TrouillonDGWRB17}. The results are shown in Figure \ref{img::noisy}.
\par According to the noisy link prediction results, we could first observe that the performance of KGE models is gradually declining as the noisy links in the training data increase. Further, the models enhanced with CausE outperform the baseline models on different benchmarks and score functions. Such experimental results show that our design is effective to counter the noise in the data set and achieve better link prediction performance.

\begin{table}
\caption{Ablation Study result on WN18RR dataset with ComplEx score.}
\label{table::ablation}
\centering
\resizebox{0.5\columnwidth}{!}{
\begin{tabular}{cc|cccc}
\toprule
\multicolumn{2}{c|}{Model}                                                                                          & MRR   & Hit@10 & Hit@3 & Hit@1 \\ \midrule
\multicolumn{2}{c|}{CausE-ComplEx}                                                                                  & 0.467 & 0.527  & 0.482 & 0.436 \\ \midrule
\multicolumn{1}{c|}{\multirow{5}{*}{\begin{tabular}[c]{@{}c@{}}$\mathcal{L}$\end{tabular}}}         & w/o $\mathcal{L}_{caus}$        & 0.458 & 0.525  & 0.479 & 0.421 \\
\multicolumn{1}{c|}{}                                                                                 & w/o $\mathcal{L}_{conf}$         & 0.453 & 0.509  & 0.467 & 0.424 \\
\multicolumn{1}{c|}{}                                                                                 & w/o $\mathcal{L}_{inter}$         & 0.427 & 0.494  & 0.452 & 0.407 \\
\multicolumn{1}{c|}{}                                                                                 & w/o $\mathcal{L}_{aux1}$         & 0.454 & 0.508  & 0.466 & 0.426 \\
\multicolumn{1}{c|}{}                                                                                 & w/o $\mathcal{L}_{aux2}$         & 0.446 & 0.497  & 0.460 & 0.419 \\ \midrule
\multicolumn{1}{c|}{\multirow{3}{*}{\begin{tabular}[c]{@{}c@{}}$\Phi$\end{tabular}}} & subtraction & 0.454 & 0.507  & 0.464 & 0.426 \\
\multicolumn{1}{c|}{}                                                                                 & multiple    & 0.439 & 0.494  & 0.454 & 0.409 \\
\multicolumn{1}{c|}{}                                                                                 & concatenation      & 0.433 & 0.482  & 0.442 & 0.409\\
\bottomrule
\end{tabular}}
\end{table}
\subsection{Ablation Study (RQ3)}
To explore the \textbf{RQ3}, we conduct ablation studies on different components of CausE in this section. We mainly verify the effectiveness and necessity of module design from two aspects.
\par First, we remove each of the five training objectives and conduct link prediction experiments. Secondly, we validate the effectiveness of the intervention operator by replacing the addition operation $\Phi$ with other common operators.

\par Our ablation studies are conducted in the mentioned settings with ComplEx score  and WN18RR dataset, while keeping other hyper-parameters same.  The results are shown in \ref{table::ablation}. The experiment results show that all five parts of the training objective are of great significance, as the model performs worse when any of them is removed.
The performance of the model degrades most when $\mathcal{L}_{inter}$ is removed. Hence, the results emphasize that causal intervention plays a very important role in CausE. Meanwhile, when the intervention operator is changed to other settings, the performance of the model has also decreased. Thus, we could conclude that the addition operation is a pretty good choice, as it is simple but effective enough.
\subsection{Visualization}
\begin{figure}[h]
    \centering
    \includegraphics[width=0.9\linewidth]{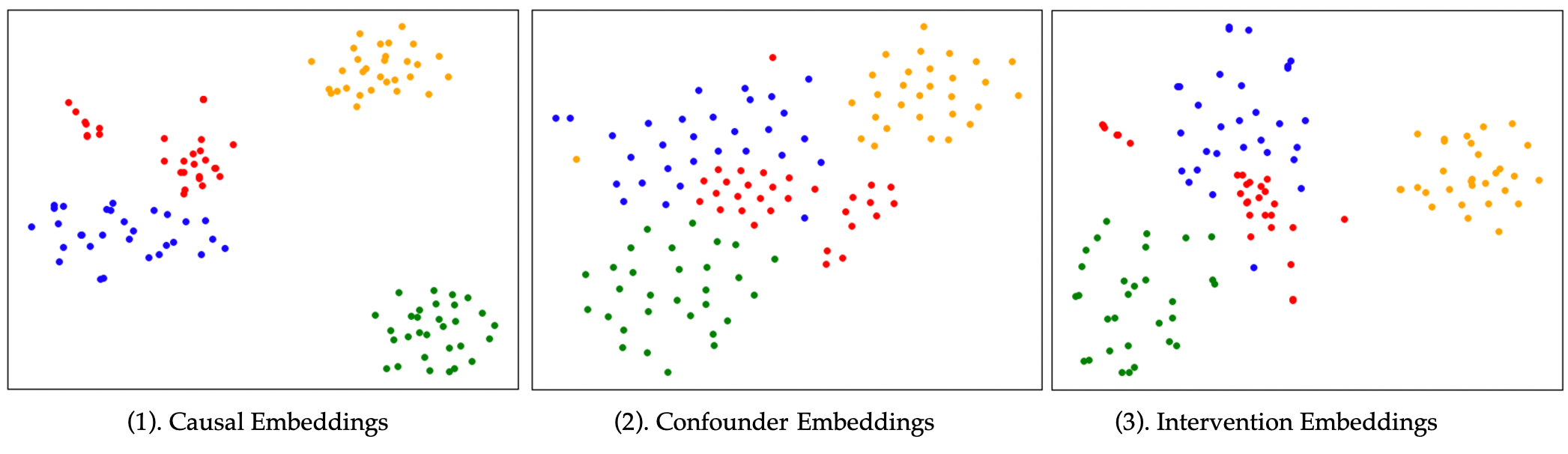}
    \caption{Embedding visualization results with t-SNE, we assign different colors for the entities with different types.}
    \label{image::vis}
\end{figure}
To answer \textbf{RQ4} and to illustrate the effectiveness of CausE intuitively , we selected entities with several different types and visualize their embeddings with t-SNE, which is shown in Figure \ref{image::vis}. We can find that the causal embedding distribution of different types can be clearly distinguished, while the confounder embedding are relatively mixed and closer together. The distribution  of the intervention embeddings which could represent the original embeddings without disentanglement  lies between the two. This shows that our approach make causal embeddings learn more distinguishable and achieve the designed goal.
\section{Conclusion}
In this paper, we emphasis that learning correlation in knowledge graph embedding models might mislead the models to make wrong predictions. We resort to causal inference and propose the new paradigm of knowledge graph embedding. Further, we propose a novel framework called CausE to enhance the knowledge graph embedding models. CausE would disentangle the causal and confounder features to different embeddings and train those embeddings guided by the causal intervention. Comprehensive experiments demonstrate that CausE could outperform the baseline methods achieve new state-of-the-art results. In the future, we plan to introduce more causality theory into knowledge graph embeddings and we attempt to apply the causal theory in more complex scenarios such as multi-modal knowledge graphs, and temperal knowledge graphs.

\section*{Acknowledge}
This work is funded by Zhejiang Provincial Natural Science Foundation of China (No. LQ23F020017), Yongjiang Talent Introduction Programme (2022A-238-G), and NSFC91846204/U19B2027.

%
%
%

\bibliography{ref.bib}
\bibliographystyle{splncs04}

\end{document}